\documentclass{article}

     \PassOptionsToPackage{numbers, compress}{natbib}

    \usepackage[final]{neurips_2023}

\usepackage[utf8]{inputenc} 
\usepackage[T1]{fontenc}    
\usepackage{hyperref}       
\usepackage{url}            
\usepackage{booktabs}       
\usepackage{amsfonts}       
\usepackage{nicefrac}       
\usepackage{microtype}      
\usepackage{xcolor}         

\usepackage[utf8]{inputenc}
\usepackage{lipsum}   
\usepackage{marvosym}
\usepackage{adjustbox}
\usepackage{graphicx} 
\usepackage{tikz}          
\usepackage{booktabs}      
\usepackage{amsmath} 
\usepackage{amssymb}  
\usepackage{color}
\usepackage{float}
\usepackage{caption}
\usepackage{subcaption}
\usepackage{hvfloat}
\usepackage{mathtools}
\usepackage{multirow}
\usepackage[title]{appendix}
\usepackage[normalem]{ulem}
\usepackage{xcolor}
\usepackage{colortbl}
\usepackage{epigraph}
\usepackage{bbm}
\usepackage{comment}
\usepackage{etoolbox}
\usepackage{rotating}
\usepackage{tabularx}
\usepackage{empheq}
\usepackage{pifont}
\usepackage{longtable}
\usepackage{makecell}
\usepackage{setspace}
\usepackage{lineno}
\usepackage[normalem]{ulem}

\usepackage[numbers]{natbib} 

\usepackage{amsthm}
\newtheorem{definition}{Definition}[section]{\bf}{\rm}
{\bf}{\it}
{\bf}{\it}
{\bf}{\rm}
{\bf}{\rm}
{\bf}{\rm}
{\bf}{\rm}
{\bf}{\rm}
{\bf}{\rm}
{\bf}{\rm}
{\bf}{\rm}
{\bf}{\rm}
{\bf}{\rm}
{\bf}{\rm}
{\bf}{\rm}
{\bf}{\rm}
\definecolor{Mycolor}{HTML}{00F9DE}

\DeclareMathOperator*{\argmin}{arg\,min}         

\newcolumntype{L}{>{\centering\arraybackslash}m{ 0.8\textwidth }}

\makeatletter
\providecommand{\subtitle}[1]{
  \apptocmd{\@title}{\par {\large #1 \par}}{}{}
}
\makeatother

\title{Privacy Measurement in Tabular Synthetic Data: State of the Art and Future Research Directions}

\author{
  Alexander T.P. Boudewijn \\
  Aindo\\
  AREA Science Park, Trieste, Italy\\
  \texttt{alexander@aindo.com} 
  \And
   Andrea Filippo Ferraris\\
   University of Turin, Italy;\\
   Data Valley consulting srl \\
  \texttt{andreafilippo.ferraris@unito.it}
  \And
  Daniele Panfilo \\
  Aindo \\
  AREA Science Park, Trieste, Italy \\
  \texttt{daniele@aindo.com} 
  \And
  Vanessa Cocca \\
  Data Valley consulting srl \\
  \texttt{v.cocca@datavalley.it} 
  \And
  Sabrina Zinutti \\
  Aindo \\
  AREA Science Park, Trieste, Italy \\
  \texttt{sabrina@aindo.com} 
  \And
  Karel De Schepper\\
  Leuven, Belgium\\
  \And
  Carlo Rossi Chauvenet \\
  Bocconi University, Italy \\
  \texttt{carlo.rossi@unibocconi.it}}

\begin{document}

\maketitle

\begin{abstract}
     Synthetic data (SD) have garnered attention as a privacy enhancing technology. Unfortunately, there is no standard for quantifying their degree of privacy protection. In this paper, we discuss proposed quantification approaches. This contributes to the development of SD privacy standards; stimulates multi-disciplinary discussion; and helps SD researchers make informed modeling and evaluation decisions.
\end{abstract}

\section{Introduction and Relation to Prior Research}
Synthetic data (SD) is rapidly gaining recognition as a privacy enhancing technology (PET)~\cite{TechSonar, Gartner}, preserving analytic value whilst removing links to real individuals. The plethora of approaches makes SDset's degree of individuals' privacy protection is hard to assess. In this paper, we codify common technical assessment frameworks for individual's privacy in SDsets. This raises interdisciplinary awareness of privacy in SD and helps SD researchers make informed modeling and assessment choices.

Several surveys mention privacy protection as an SD use case, but do not cover its assessment in a detailed manner~\cite{vanbreugel2023privacy,jordon2022synthetic, SDDP2}. Reviews of privacy in AI fail to metion SD~\cite{decristofaro, Aggarwal2008}. Surveys, reviews, and experimental comparisons of SD techniques provide little consideration of privacy metrics~\cite{HERNANDEZ202228, Kartik, SDDP, SDDP2, NIST, Figueira, Hittmeir, Raab}. Legal analyses of SD are scarce and do not cover quantitative, case-by-case privacy assessment methods~\cite{Bellovin18, Cesar}.


\section{Definitions and Notation}\label{sec:def}
To the best of our knowledge, there is no widely accepted definition of SD. Following Jordon et al.~\cite{jordon2022synthetic}, we propose Definition~\ref{def:SD}.

\begin{definition}\label{def:SD} \emph{(Synthetic data,~\cite{jordon2022synthetic})}
\textbf{Synthetic data} (SD) is data that has been generated using a purpose-built mathematical model or algorithm (the ``\textbf{generator}''), with the aim of solving a (set of) data science task(s).
\end{definition}

The \emph{generator} can be inferred through deep learning, (e.g. Generative Adversarial Networks (GANs)~\cite{orig_Gan, GAN, orig_GAN_tab, orig_CTGAN_tabular}; Variational Autoencoders (VAEs)~\cite{orig_VAE, orig_b_VAE, huang2018introvae, PanfIEEE}); agent-based and mathematical modelings~\cite{Agentbased, probabilistic}; autoregressive approaches through traditional AI, e.g. decision tree learning~\cite{synthpop, AR}; diffusion models~\cite{Diffusion,kotelnikov2022tabddpm}; nearest neighbor-based methods~\cite{Simulants, avatar}; Bayesian networks~\cite{bayes}; clustering~\cite{clustering}; and large language models~\cite{language}.

We let $D$ denote a database describing \textbf{data subjects} through attributes $A(D)$. Rows $d\in D$ are $|A(D)|$-tuples with a value $v(d,a)$ for each attribute $a\in A(D)$. Attribute $a\in A(D)$ is \textbf{categorical} if its domain is finite and \textbf{numeric} if its domain is a subset of $\mathbb{R}$. We use the terms \textbf{row} and \textbf{record} interchangeably. We denote by $\mathcal{G}$ a generator, and by $\hat{D}\sim\mathcal{G}(D)$ denote that synthetic dataset $\hat{D}$ was obtained from generator $\mathcal{G}$ trained on $D$. \textbf{Seed-based} generators are a subclass of generators that produce one unique synthetic record, denoted $\mathcal{G}(d)$ for every given real record $d$ (the \textbf{seed}). This is opposed to most models (e.g. GANs, VAEs) that represent the overall properties of datasets probabillistically, and then produce synthetic data by randomly sampling from the obtained distribution, breaking the one-to-one correspondence between real and synthetic records.

\section{Synthetic Data Privacy Risks}\label{sec:risks}
Three key risks identified by the WP 29 ~\cite{wp29anon}, act as benchmark for a proper anonymization, namely: \emph{Singling Out} (isolating records), \emph{Linkability} (linking records concerning the same data subject in one or more datasets), and \emph{Inference} (deducing, with significant probability, the value of an attribute). Privacy risks in SD can be a consequence of various factors. The most important ones are detailed below.

\textbf{Model and data properties.} Improperly trained Generators may overfit, memorizing and reproducing fixed patterns rather than inferring stochastically~\cite{Mitchell, memorization}.  Records that emerge in isolation, with little variability around their attribute values are difficult to generalize. As such, datasets with outliers; sparse datasets; and datasets with underrepresented strata are more at risk of memorization than more homogeneous sets~\cite{KDP, task}. By their natures, such sets also have large singling-out susceptibility.

\textbf{The approach to data synthesis.}  Most generators represent overall datasets stochastically, and obtain synthetic records by random sampling. This removes links between real data subjects and synthetic records. However, some methods (e.g.~\cite{avatar, Simulants}) create one specific synthetic record for each real record. This poses greater risk, as the link between data and data subject is maintained.

GANs may infer the minimal information needed to deceive the discriminator, failing to capture the nuances and variability of real data (mode collapse~\cite{Figueira, modeCollapse, modeCollapse2}). The SD then resembles a small selection of real data subjects well, but not the population as a whole. The SD becomes ``cluttered'' around specific real records, leaking information about them (see Appendix~\ref{app:risk}, Figure~\ref{fig:mode collapse}).

\begin{figure}[H]
       \centering
        \includegraphics[height=0.2\textheight]{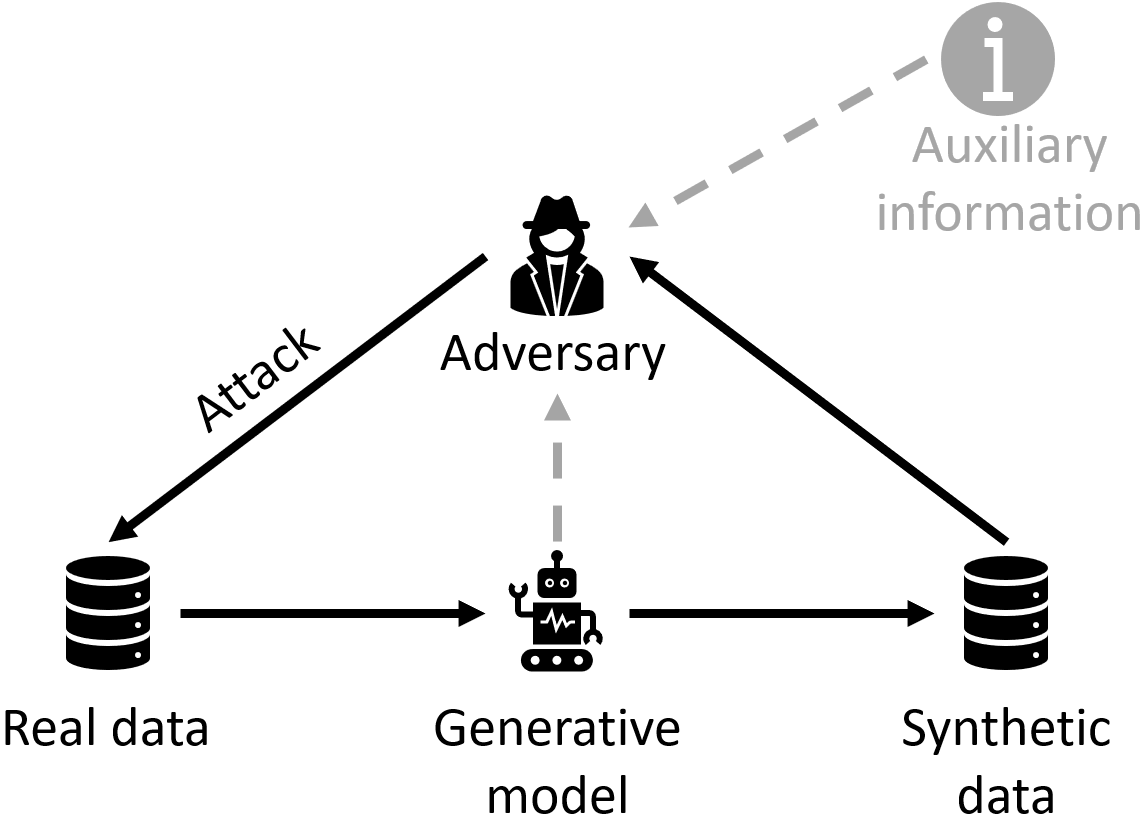}
        \caption{Visualization of an attack. Dotted lines indicate that the adversary may or may not leverage information sources, depending on the threat model: the adversary may use auxiliary information and/or information about the generative model (no, black, uncertain, or white box).}
        \label{fig:base1}
    \end{figure}

\textbf{The threat model.} A threat model is the information leveraged by an adversary besides the SD (see Figure~\ref{fig:base1}). They can be: 1) \emph{No box}: the adversary accesses the SD only. 2) \emph{Black box}: the adversary also has limited generator access (e.g. no access to the model class or parameters, but access to the model's input-output relation). 3) White box: the adversary has full generator access (model class and parameters). 4) \emph{Uncertain box}~\cite{TAPAS}: the adversary has stochastic model knowledge (model class and knowledge that parameters stem from given probability distributions). 5) Any of the aforementioned, along with \emph{auxiliary information}; in the context of SD formalized through Definition~\ref{def:auxil}.

\begin{definition}\label{def:auxil}
    Let $D$ be a dataset with attributes $A(D)$. An adversary has \textbf{auxiliary information} if they know the values of some subset $A'\subseteq A(D)$ of attributes of some subset $D'\subseteq D$ of records. 
\end{definition}

\section{Mathematical Privacy Properties}\label{sec:math}
\subsection{Differential Privacy}
Differential privacy (DP)~\cite{Dwork2014dif} is a property of information-releasing systems. A DP system does not release data directly, but a derivative obtained through processing. The system is considered DP if the released information does not change significantly when a single record is removed from the database. DP is formally defined in Definition~\ref{def:DP}.    

\begin{definition}\label{def:DP}\emph{(Differential Privacy,~\cite{Dwork2014dif})}
A randomized algorithm $\mathcal{M}$ is $(\varepsilon,\delta)$-\textbf{differentially private} ($(\varepsilon,\delta)$-DP) if for all $S\subseteq A(P)$:
\begin{equation}\label{eq:DP}
        \mathbb{P}\left[\mathcal{M}(D)\in S\right]\leq e^{\varepsilon}\cdot\mathbb{P}\left[\mathcal{M}(D')\in S\right]+\delta
\end{equation}
for all databases $D, D'$ such that $\exists d\in D: D'=D\setminus\left\{d\right\}$.
\end{definition}

\begin{figure}[H]
        \centering
        \includegraphics[height=0.25\textheight]{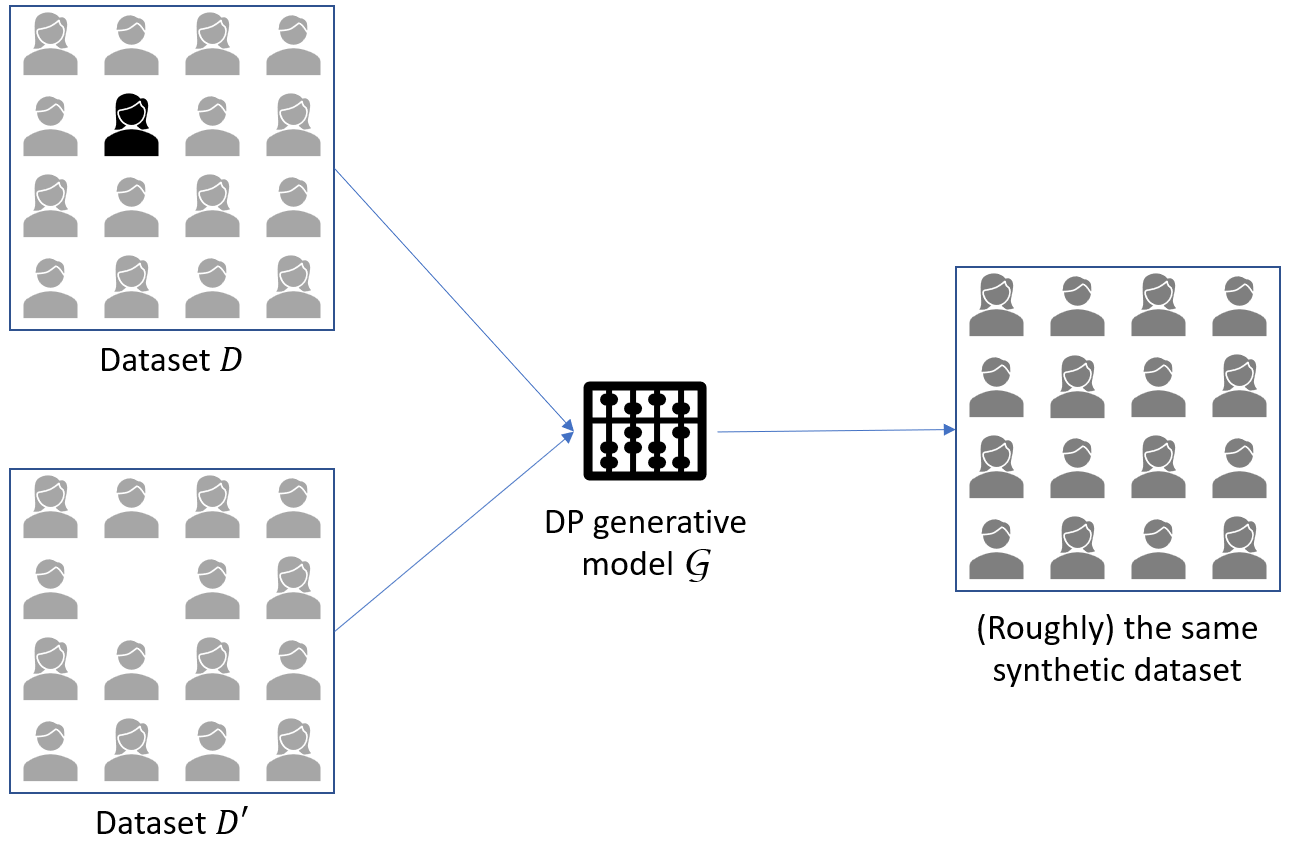}
        \caption{Differential privacy in synthetic data generative models: the SD released by generator $\mathcal{G}$ does not alter significantly if any single individual $d$ (indicated in black) is removed from the training data. This protects the data subject with record $d$, as it makes it impossible to pinpoint information to them.}
        \label{fig:DP}
    \end{figure}

Generators are data releasing systems and can thus be DP: suppose we have two real datasets $D$ and $D'$ with $D'=D\setminus\left\{d\right\}$. Then generator~$\mathcal{G}$ is DP if a data controller with access to $\hat{D}\sim\mathcal{G}$ cannot infer whether $\mathcal{G}$ was trained on $D$ or $D'$ (Appendix~\ref{app:DP}, Figure~\ref{fig:DP}). Appendix~\ref{ap:DPmech} details approaches to train generators with built-in mechanisms to guarantee output data is DP. Importantly, in this context, DP is a \emph{property of generators}, and not of the synthetic data they may produce.
       
\subsection{\texorpdfstring{$k$}{k}-Anonymity}
Privacy risks persist even if identifying attributes like names are removed: combinations of attribute values may still single out an individual. The concept of $k$-anonymity was introduced to avoid thereby incurred risks~\cite{wp29anon, sweeney2002k, Kano, KANN_Comp}. A dataset is $k$-{\bf anonymous} if at least $k$ individuals share each combination of attribute values. Further restrictions ($l$-diversity~\cite{ldiv}; $t$-closeness~\cite{tclose}; $(\alpha,k)$-anonymity~\cite{akanonymity}) offer additional protection.

Synthetic data based on autoregressive models can incorporate $k$-anonymity directly in the generation process~\cite{synthpop}. For example, in data generated by decision trees, pruning can guarantee that each combination of attribute values is sampled at least $k$ times in mathematical expectation~\cite{Rankin}. Unlike DP, $k$-anonymity is a property of deidentified or synthetic datasets, not the algorithms producing them.

\subsection{Plausible Deniability}
A degree of plausible deniability is inherent in synthetic datasets, as their records do not pertain to real data subjects. Two approaches have emerged to formalize the notion of plausible deniability~\cite{plaus, plaus2}, of which one is most relevant to (seed-based) synthetic data.

    \begin{definition}\label{def:PD1}
        \emph{(Plausible deniability (PD),~\cite{plaus})} Let $D$ be a dataset and let $\mathcal{G}$ be a generator that converts any real individual record $d\in D$ into a corresponding synthetic record $\hat{d}=\mathcal{G}(d)$. For any dataset $D$ with $|D|>k$, and any record $\hat{d}$ such that $\hat{d}=\mathcal{G}(d_1)$ for $d_1\in D$, we say that $\hat{d}$ is releasable with $(k,\gamma)$-\textbf{plausible deniability}, if there exist at least $k-1$ distinct records $d_2,...,d_k\in D\setminus\left\{d_1\right\}$ such that for all $i,j\in\left\{1,2,...,k\right\}$:
        \begin{equation}\label{eq:PD1}
        \gamma^{-1}\leq \frac{\mathbb{P}\left[\hat{d}=\mathcal{G}(d_i)\right]}{\mathbb{P}\left[\hat{d}=\mathcal{G}(d_j)\right]}\leq\gamma
        \end{equation}
    \end{definition}

\noindent Intuitively put, a generator producing synthetic records from a particular seeds has PD if, for each synthetic record generated from a specific seed, $k$ other seeds could have resulted in \emph{roughly the same} (quantified through $\gamma$) synthetic record. Like DP and unlike $k$-anonymity, PD is therefore a property of (seed-based) generators, though it shares intuition with both other properties.

\section{Statistical Privacy Indicators}\label{sec:indicators}
\subsection{Identical Records, Distances, and Nearest Neighbors}\label{sec:disteval}
Most indicators quantify how many synthetic records are identical, or suspiciously similar to particular real records. Unlike DP and PD, these indicators measure \emph{properties of synthetic datasets}, not their generators. The proportion of synthetic records that coincide with real records is referred to as the \emph{identical match share} (IMS)~\cite{Ebert, Platzer, 5g}. The IMS is therefore generalized to similarity metrics, and further to Nearest neighbor (NN)-based methods. The latter two can be classified based on the properties detailed below. Table~\ref{tab:indicators} of Appendix~\ref{app:dist} classifies approaches along these properties.

\textbf{Similarity metrics.} Table~\ref{tab:distMet} in Appendix~\ref{app:dist} contains an overview of commonly invoked measures.

\textbf{Metric evaluation.} A complicating factor in evaluating similarity metrics in structured datasets is the multitude of datatypes. The following approaches exist to do so: 1) \emph{binning numeric attributes}; treating them as categorical; and using a metric for categorical values. 2) \emph{Aggregation of multiple metrics}, applying one metric per type and integrating the results. 3) \emph{Ignoring attributes}, for instance by considering only numerical attributes with a metric appropriate for them. 4) \emph{Evaluating distances in embedding spaces}, in which all information is preserved, but represented in normalized, numeric attributes, e.g. through t-SNE~\cite{tsne}, discriminant analysis~\cite{disca}, factor analysis~\cite{Saporta}, or representation learning~\cite{shenkar2021anomaly}. 

\textbf{Evaluated distances.} For a given synthetic record~$\hat{d}\in \hat{D}$, we can find its closest real record~$d\in D$. We call the distance between these records the \emph{synthetic to real distance (SRD) of~$\hat{d}$}, denoted by~$\texttt{SRD}(\hat{d})$ (see equation~\eqref{eq:dsr}, where~$\texttt{Dist}$ is some similarity metric).
\begin{equation}\label{eq:dsr}
    \texttt{SRD}(\hat{d}) := \min\limits_{d\in D}\texttt{Dist}(\hat{d},d)~~~~\forall\hat{d}\in\hat{D}
\end{equation}
\noindent In an analogous fashion, the smallest synthetic to synthetic ($\texttt{SSD}$), real to synthetic ($\texttt{RSD}$), and real to real distance ($\texttt{RRD}$) can be defined. These are all visualized in Figure~\ref{fig:dist} of Appendix~\ref{app:dist}. 


\textbf{Use of holdout sets.} To compute the RRD, the real data $D$ can be partitioned into two subsets $D_1$ and $D_2$. For a real record $d_1\in D_1$, the RRD is then the smallest distance to any record $d_2\in D_2$, as in equation~\eqref{eq:RRD}.  This ``holdout set'' provides a baseline for comparing SD against~\cite{Platzer}. 
\begin{equation}\label{eq:RRD}
    \texttt{RRD}(d_1):=\min\limits_{d_2\in D_2}\texttt{Dist}(d_1,d_2)~~~~\forall d_1\in D_1
\end{equation}

\textbf{Statistics.}  The \emph{Distance to closest record} (DCR) compares the SRD and RRD distributions. Real data subjects may be at risk if, for some synthetic record $\hat{d}$, we have $\texttt{SRD}(\hat{d})<\texttt{RRD}(d^*)$, with $d^*:=\argmin_{d\in D}\texttt{Dist}(\hat{d},d)$. The DCR is sensitive to realistically replicated outliers, as they have large RRDs (see Appendix~\ref{app:dist}, Figure~\ref{fig:SRD}). Risks are expressed statistically through proportions~\cite{PanfIEEE, PhD_Dani} and medians, means and standard deviations of ``suspiciously close'' synthetic records, with~\cite{weldon2021generation,Platzer,NeurIPS,PanfIEEE} or without~\cite{an2023distributional, Diffusion, kotelnikov2022tabddpm} using a hold-out set. Small percentiles are also often invoked~\cite{NeurIPS, Ebert, zhao2021ctabgan, avatar}. E.g., Mami et al.~\cite{NeurIPS} compute the proportion $P$ of synthetic records that closer to real records than the smalles 5\% of RRDs, using a holdout set. 

Panfilo et al.~\cite{PanfIEEE, PhD_Dani} use an \emph{inferential statistical test} to assess whether the SRD and RRD stem from the same distribution. Yale et al.~\cite{Yale, Yale2, Yale3} introduced the \emph{adversarial accuracy} and \emph{Privacy loss}, including the SSD and RSD for a baseline. Some distance-based indicators are for seed-based SD only: \emph{distance-based record linkage}~\cite{DBRL, HERRANZ201578, DBRLSynth}; the \emph{hidden rate}~\cite{avatar}; and \emph{local cloaking}~\cite{avatar}.

\subsection{Other Statistical Indicators}
Taub et al.\cite{Taub2019CreatingTB} introduce the \emph{targeted correct attribution probability} (TCAP) indicator. This TCAP is essentially an indicator of parameter inference attack success rates. It quantifies the frequency with which synthetic parameter values correspond to real ones in $l$-diverse equivalence classes. Emam et al.~\cite{Emam} derive a related probabilistic approach to quantify the risks of the WP29 attacks, using real holdout sets as baselines. Esteban et al.~\cite{esteban2017realvalued} and Rashidian et al.~\cite{smooth} propose using the maximum mean discrepancy (MMD) as a privacy metric, inferentially testing whether the generator overfits.

\section{Computer Scientific Experimental Privacy Assessment}\label{sec:attack}

Computer scientific privacy assessment is the deliberate conducting of SD-informed privacy attacks, and the measurement of their effectiveness, to quantify SD's degree of protection. Attack frameworks are classified in Table~\ref{tab:attack_comp} (Appendix~\ref{app:attack}), based on threat models and the factors outlined below. 

The use of specific threat models is an important innovation of the computer scientific approach. Mathematical properties and statistical indicators pertain only to either generators or synthetic data (but not both). Computer scientific attacks, on the other hand, allow for flexibility in modeling how much knowledge an adversary may have about generators.

\subsection{Attack Frameworks}

\textbf{Vulnerable Record Discovery (VRD).} Some methods conduct attacks by identifying seemingly vulnerable synthetic records. Giomi et al.~\cite{Anonymeter} propose looking for synthetic records with unique (combinations of) attribute values. Singling out attacks are then claims that these records are also unique real records. Carlini et al.~\cite{carlini2019secret} study the extent to which generators overfit, quantifying the likelihood of synthetic records being memorized secrets. 

\textbf{Adversarial Machine Learning.} Model inversion, membership inference attacks (MIAs), and shadow modeling (``model stealing'') compromise confidentiality without technological system misuse~\cite{adversarial, adversarial2, rigaki2021survey}. In a MIA, an adversary infers whether a given target record was in the training dataset of a given ML model. This can be through classifier models~\cite{choi2018generating, Kuppa, Park_2018, vanbreugel2023membership, Yale, oprisanu2022utility, Goncalves, groundhog}.  A \emph{shadow model} (SM) is constructed by an adversary to mimic a given model. SMs may mimic a given generator  to conduct MIAs~\cite{Park_2018, TAPAS, Kuppa, knock}: data is provided to  the SM and the real model. By comparing their outputs, the adversary determines whether a given record was in the training set of the real model~\cite{membership, groundhog, vanbreugel2023membership, oprisanu2022utility}. Shadow modeling requires at least a black box threat model.

\textbf{Combined approaches.} Recent approaches use VRD to make informed decisions for potential targets in membership inference attacks, reducing the computational burden~\cite{Diffusion, vanbreugel2023membership, meeus2023achilles, MIAnonOutlier}.

\subsection{Attack Mechanisms}
\textbf{Nearest Neighbors (NN).} Suppose an adversary has auxiliary information about a target, but does not know the value of one of its attributes. They may then assign the missing value based on the target's $k$ synthetic NNs (see Appendix~\ref{app:attack}, Figure~\ref{fig:NNattack})~\cite{choi2018generating, Goncalves, Yale3, Anonymeter, TAPAS, HideNSeek, HideNSeek2}. Experimentally, small $k$ values perform well, particularly $k=1$~\cite{Goncalves, choi2018generating, Anonymeter, HideNSeek2}. More auxiliary information means better attacks: Experimentally, access to one extra attribute of auxiliary information roughly increases accuracy by 30\% across datasets and generators~\cite{Goncalves}. 

\textbf{Machine Learning (ML).} Techniques from ML can guide the attack process. For instance, classifiers can be trained to re-identify real data subjects~\cite{membership, groundhog, Park_2018, vanbreugel2023membership}. Classifiers and regression models can also be used to estimate parameter values of real records based on synthetic ones~\cite{choi2018generating, Goncalves}.

\textbf{Information Theory (IT).} IT concepts like Shannon entropy~\cite{Shannon} and mutual information can be used to identify the degree to which records in an anonymized dataset deviate from more common ones, in which case they have a higher likelihood of being memorized by an overfit generator~\cite{task, carlini2019secret, narayanan2006break, deanonymization}.

\subsection{Baselines and Effectiveness Estimation}
\textbf{Absolute Measurements.} Efficacy metrics not requiring baselines include the probability with which records can be singled out~\cite{carlini2019secret, deanonymization, narayanan2006break}; and the proportion of real records for which information can be re-identified~\cite{choi2018generating, Goncalves}. ML-based attacks can be evaluated through ML metrics (see~\cite{Mitchell}). For instance, MIAs based on classification are evaluated with ROC AUCs ~\cite{vanbreugel2023membership, Yale3, Park_2018}; F-1 scores~\cite{Kuppa, Park_2018} and Precision and recall~\cite{choi2018generating}. 

\textbf{Random baseline.} Giomi et al.~\cite{Anonymeter} propose a random baseline, to evaluate attack efficacy with SD access to that of uninformed (random) guesses (see Appendix~\ref{app:attack}, Figure~\ref{fig:base2}). For some methods, the mathematical expectation of random hypotheses is known a priori and implicitly integrated in scoring (see, e.g.~\cite{HideNSeek2}).

\begin{figure}[H]
       \centering
        \includegraphics[width=0.6\textwidth]{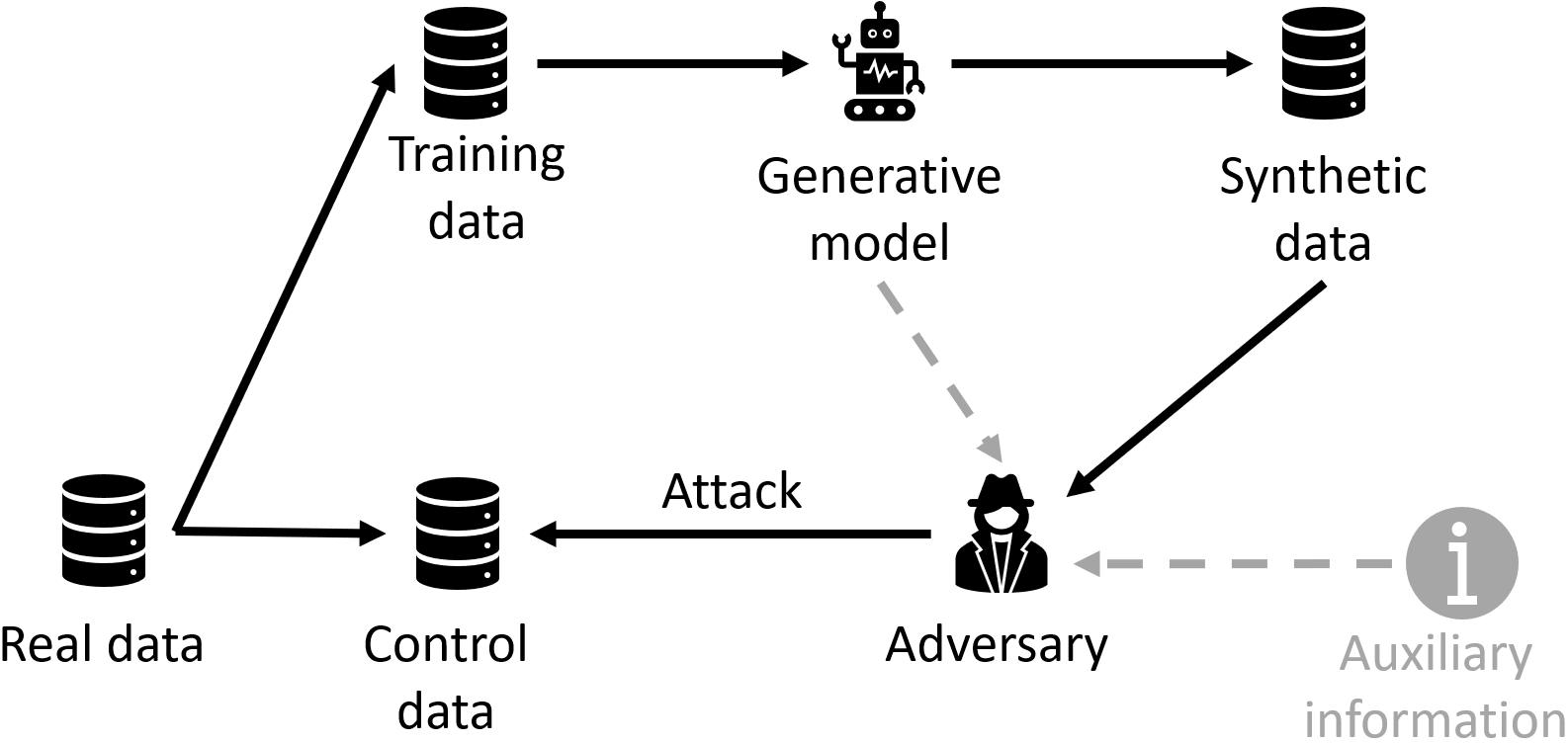}
        \caption{Control baseline: the adversary uses the available information to target a control data, not used in training the generative model}
        \label{fig:base3}
    \end{figure}

\textbf{Control baseline.} Giomi et al.~\cite{Anonymeter} propose using a \emph{control baseline}, by splitting the real data into a training set and a control set. The generator is trained using the training set, having no access to the control set. The estimated success rate of attacks on the training data (Figure~\ref{fig:base1}) is compared to that of attacks on the control data (Figure~\ref{fig:base3}). If the former is large, the SD may leak information. However, if the latter is also large, this was generic, population-level information, as opposed to specific secrets of specific data subjects (``\emph{relating to in content}'' in the legal analysis of L\'{o}pez and Elbi~\cite{Cesar}).  

\textbf{Deliberate secret insertion.} Deliberate secrets can be inserted in the training data~\cite{carlini2019secret}, or the SD after generation~\cite{Anonymeter}. Re-identifying these secrets casts light on the ease of inferring sensitive information from synthetic data. 

\subsection{Relation to WP29 Attack Types}
\textbf{Singling out.} VRD directly implements singling out attacks, identifying SD records that likely result from overfit generators (typically outliers). This works even under a no box threat model. MIA can also model singling out: the adversary quantifies the likelihood of a combination of attribute values being a unique real record.

\textbf{Linkage.} Attacks with NN as mechanism require auxiliary information. They may be interpreted as linkage attacks: the adversary has to obtain the auxiliary information from some other data source, linking it to the SD. The \emph{Anonymeter} framework~\cite{Anonymeter} and information theory-based VRD~\cite{narayanan2006break} are the only methods to explicitly model linkage attacks.

\textbf{Inference.} NN-based attacks and MIA can function as inference attacks. E.g. if the SD is used to evaluate the rare disease treatment efficacy, the training set contains patient data. If an adversary can determine that Giovanna is in the training dataset, they can infer that she has the disease. 

\section{Discussion}\label{sec:conc}
\subsection{The Assessment Frameworks}
\textbf{Mathematical privacy properties.} In DP, there is no consensus on the choice of parameters $(\varepsilon, \delta)$. Large parameter values offer weak privacy guarantees. A given~$\varepsilon$ can result in different degrees of protection for different use cases, making it hard to choose values in practice~\cite{epsilon}. Furthermore, DP SD is still susceptible to linkage and inference attacks~\cite{wp29anon, stadler0}, possibly providing a false sense of security. DP is a property of generators, not their produced data.

Achieving $k$-anonymity involves considerable information destruction~\cite{kannUtil} and is NP-hard to achieve optimally through generalization~\cite{meyerson}. In a court ruling in California, $k$-anonymity was shown to offer sufficient protection, only once the analytic utility is completely removed~\cite{Sander_VS_Cali}. This is corroborated by findings that with a combination of only fifteen parameter values, over 99\% of a population can be re-identified~\cite{Montjoye}. Unlike DP, $k$-anonymity is a property of deidentified datasets, not the methods that produce them. 

PD is only Applicable to seed-based methods, ruling out most classes of generators. Like $k$-anonymity, an individual record is considered protected if it is indistinguishable from a fixed number of other records. The difference is that PD is developed for synthetic data specifically, with ``indistinguishable'' the probability of stemming from multiple seeds. The notion of probabilistic lack of impact of an individual record also shares its intuition with DP, with which PD is closely related. To date, PD has gained little traction in practice. 

\textbf{Statistical privacy indicators.} Distance-based indicators (by far the most common indicators) are difficult to interpret. The multitude of options and involved modeling decision adds to this confusion. Evaluation of distances can be a particular difficulty, as structured data has mixed data types, for which different similarity metrics may be appropriate. Choice of similarity metric and its evaluation may have an impact on results. Statistical indicators measure properties of synthetic data, not their generators. 

\textbf{Computer scientific privacy experiments.} Deliberate attacks are most commonly MIAs. The results of such assessments provide crucial insight into data privacy. However, they often only work under threat models with considerable information. No box, no auxiliary information approaches are more rare, and typically confined to outlier detection (VRD). 

Most attack-based approaches require auxiliary information, Arguably making them linkage attacks. Most attack approaches rely on distances (nearest neighbors) or ML as an attack mechanism, so we hypothesize that distance-based indicators are very reliable predictors of their efficacy, quantifying risks in a more all-encompassing manner.

Unlike the other approaches, computer scientific experiments can leverage several threat models. This allows them to take into consideration properties of both the synthetic data and their generators.

\subsection{Relation to Synthetic Data Risks}
A core SD risk is generator memorization, particularly around outliers; in sparse or small datasets; or through mode collapse. All frameworks address this risk: mathematical properties center around uniqueness of records. DP measures the impact of individual training records, with outliers clearly having large individual impacts. Furthermore, $k$-anonymity and PD specifically foster datasets with severely limited uniqueness of individual records. 

Distance-based indicators are sensitive to outliers, as their synthetic neighbors have small SRDs, while the corresponding real outliers have large RRDs (Figure~\ref{fig:SRD}). Other statistical indicators are measures of memorization by their very nature. In computer scientific experiments, VRD deliberately seeks for outliers, while MIAs are nearly exclusively effective for outliers (see, e.g.~\cite{groundhog}).

To the best of our knowledge, no research was conducted to assess whether seed-based generators inherently pose greater risks than other generators. Intuitively, this seems evident, as they do not remove the links between (synthetic) records and real data subjects. 

\subsection{Suggestions for Future Research}
\textbf{Standardizing privacy assessment.} Synthetic data privacy is a multifaceted subject encompassing several disciplines such as mathematics, computer science, ethics, policy-making, law, and philosophy. There is a pressing need for increased interdisciplinary research to gain an inclusive understanding of synthetic data as a PET. Conventional assessment standards should be developed, so that research findings are easy to interpret, compare, and contrast. Consensus should be formed over whether privacy is a property of synthetic datasets, the generators that produce them, or some combination of both.

\textbf{Synergies between assessments.} A comparison (deductive or experimental, e.g. comparing multiple assessments on the same SDsets) between mathematical, statistical, and empirical privacy approaches would indicate consistency, and identify merits and weaknesses. For replicability, experiments should use open-source generators and publicly available datasets (e.g. from the UCI ML repository~\cite{Dua:2019}). As attacks are often distance (NN)-based, insights from indicators should be integrated in simulated attacks (e.g. involved distance metrics and their evaluation methods; distances; use of holdout set for a baseline; statistical interpretation of results).

\textbf{Outlier protection.} Following Tai et al.~\cite{KDP}, research should address outlier protection in SD, e.g. by binning and aggregating attribute values (cf. Section~\ref{sec:disteval} on metric evaluation); through innovation; or by invoking other PETs. Outlier detection methods (e.g.~\cite{Chandola, Pang, Ruff_2021}) can be used for VRD.

\textbf{Incorporating privacy into generators.} While DP is incorporated in various generators, this is not true for privacy metrics and empirical privacy approaches. Future research should focus on incorporating the latter two, for instance by incorporating metrics in loss functions, or through combinatorial optimization. Combining SD with outlier-protecting PETs should be considered.

\textbf{Assessment for advanced data formats.} Most covered approaches to privacy assessment in structured data were developed for data contained in a single table. More research is required to assess privacy in relational datasets, with information contained in multiple, interconnected tables. So-called ``\emph{profiling attacks}'' re-identify subjects not by their literal records, but by latent behavioral patterns~\cite{Profiling}. Such attacks may play a more considerable role in the context of relational databases.

\textbf{Distribution-level confidentiality.} The outlined frameworks are developed to assess the upholding individuals' right to privacy. In practice, properties of datasets as a whole may additionally be confidential. They may for instance be trade secrets (e.g. the total number of annual transactions of a financial institution). The reader is referred to~\cite{zhang2021attribute, lin2023summary, suri2022dissecting} for contemporary assessment frameworks of confidentiality on the level of overall dataset properties.

\bibliographystyle{ieeetr}
\bibliography{References}

\appendix 
\section{Synthetic Data Risks}\label{app:risk}

    \begin{figure}[H]
       \centering
        \includegraphics[height=0.2\textheight]{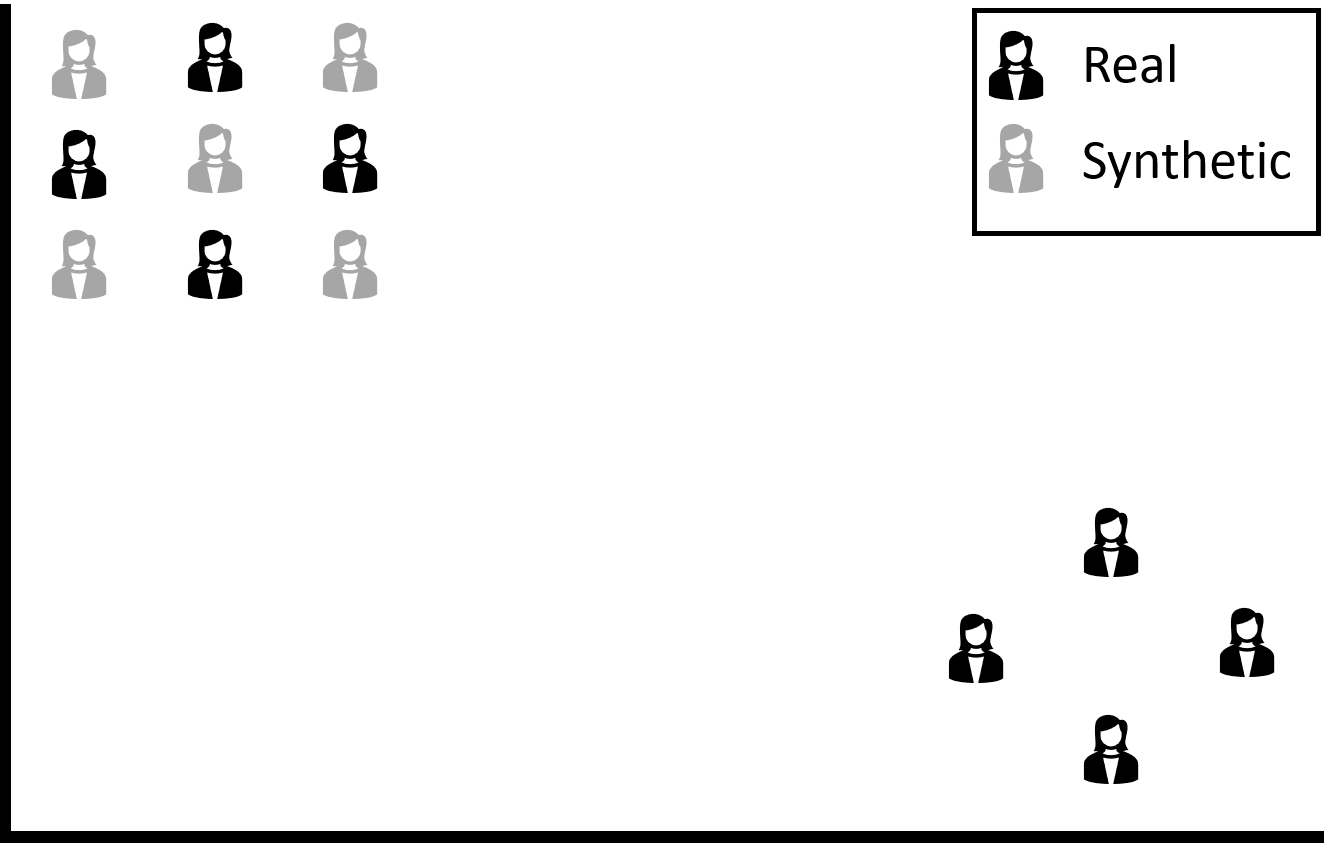}
        \caption{Mode collapse: the real records (visualized in black) are varied, and represented in two regions of the plane (top-left; bottom-right). The generator only learns to replicate synthetic records (visualized in grey) in the top-left group: merely inferring their patterns is sufficient to deceive the discriminator.}
        \label{fig:mode collapse}
    \end{figure}

\section{Differentially Privacy for SD}\label{app:DP}
\subsection{Generators as Information Release Systems}

\subsection{Overview of SD Methods with Built-In DP Mechanisms}~\label{ap:DPmech}
Table~\ref{tab:DP} contains an overview of methods for generating SD, in which DP is incorporated directly. The resulting SD, when released, then automatically guarantees DP for the real dataset. Rosenblatt et al.~\cite{DPGAN_Survey} compare the performances of three approaches to DP-guaranteed SD generation (DP-GAN, PATE-GAN, MWEM, see below), concluding that PATE-GAN has the best utility for tabular data in ML applications.

    \begin{table}[H]
    \centering
    \resizebox{\textwidth}{!}{%
    \begin{tabular}{llll}
        \toprule
    Method & SD technology & DP integration method & Reference(s) \\
     \midrule
     Differential privacy GAN  & GAN         &  Noise addition during training & \cite{Abadi_2016, DPGAN}\\
      ~~(DP-GAN)& & &~~\cite{Fang, DPGAN5} \\
      &&&\\

     Federated learning GAN  & GAN & Multiple clients train part of the & \cite{Xin}\\        ~~(FL-GAN)          &     & ~~ GAN (see~\cite{FL}) on non-overlapping, & \\
                                     &     &~~ noise-added datasets & \\
                                     &&&\\
     PATE-GAN & GAN & Multiple discriminators are trained,  &\cite{yoon2018pategan} \\
              &     & ~~ each on different subsets of data.\\
              &     & ~~ Their classifications (real or& \\
              &     & ~~ synthetic) are aggregated. &\\
              &     & ~~ noise is added during the  agg-& \\
              &     & ~~ regation process (private agg- & \\
              &     & ~~ regation of teacher ensembles, &\\ &&~~ i.e. PATE, see~\cite{PATE}) & \\
     &&&\\
     Dual adversarial & VAE, GAN & Noise addition during training & \cite{DPloss}\\
     ~~autoencoders (DAAE) &&&\\
     &&&\\
     Private-PGM & Bayesian network & Sample marginals of the real data; & \cite{RDP1} \\
                 &                  & ~~ infer high-dimensional distribution& \\
                 &                  & ~~ from these marginals through a & \\
                 &                  & ~~ probabilistic graphical model (PGM,& \\
                 &                  & ~~ see~\cite{mckenna2019graphicalmodel}); add noise during&\\
                 &                  & ~~ sampling. & \\
    &&&\\
    Simulants    & Nearest Neighbor &Noise addition during training & \cite{Simulants}\\
    &&&\\
    Multiple-level clustering  & Clustering &Noise addition during training&\cite{clustering}\\
    ~~generator (MC-GEN) & & & \\
    &&& \\
    Multiplicative Weights & Optimization & Noise addition & \cite{hardt2012simple}\\
    ~~Exponential Mechanism &&&\\ 
    ~~(MWEM) &&&\\
         
    \bottomrule
    \end{tabular}%
    }
    \caption{Overview of generative models with built-in differential privacy mechanism}
    \label{tab:DP}
    \end{table}

\section{Statistical Privacy Indicators}\label{app:dist}

    \begin{table}[H]
    \centering
    \resizebox{\textwidth}{!}{%
    \begin{tabular}{lll}
        \toprule
    Name & Computation & Remark \\
     \midrule
     $\mathcal{L}_1$-distance & $\texttt{Dist}_{\mathcal{L}_1}(d, d')=\sum_{a\in A(D)} |v(d,a)-v(d',a)|$ & Numeric attributes;\\
     && ~~ also known as \emph{Manhattan distance}\\
     &&\\
     
     Euclidean distance & $\texttt{Dist}_{E}(d, d')=\sqrt{\sum_{a\in A(D)} (v(d,a) - v(d',a))^2}$ & Numeric attributes; \\
     && \\

     Hamming distance & $\texttt{Dist}_{H}(d, d')=|\left\{a\in A(D): v(d, a) \neq v(d', a)\right\}|$ & Categorical attributes;\\
     & & \\

     Cosine similarity & $\texttt{Dist}_{C}(d,d') = \frac{\sum_{a\in A(D)} v(d,a) v(d',a)}{\sqrt{\sum_{a\in A(D)} v(d,a)^2\cdot \sum_{a\in A(D)} v(d',a)^2}} $& Numeric attributes;\\
                       &  & ~~Technically not a distance metric\\
                       &&\\

    Manhalobis distance & $\texttt{Dist}_M(d,d') = \sqrt{(d - d')S^{-1}(d-d')^T}$ &  For $S$ the covariance matrix of\\
    &&~~real and synthetic distributions\\
    &&~~Categorical attributes;\\
    &&~~Generalization of Euclidean distance\\
    &&~~taking correlation into account.\\
    &&\\
    Gower distance &Hamming distance for categorical attributes& An aggregation of two metrics \\
                   & ~~$+\mathcal{L}_1$ distance for numerical attributes&\\
         
    \bottomrule
    \end{tabular}%
    }
    \caption{Common distance and similarity metrics in synthetic data privacy assessment}
    \label{tab:distMet}
    \end{table}

\begin{figure}[H]
    \centering
    \begin{subfigure}{0.46\linewidth}
        \centering
         \includegraphics[width=\linewidth]{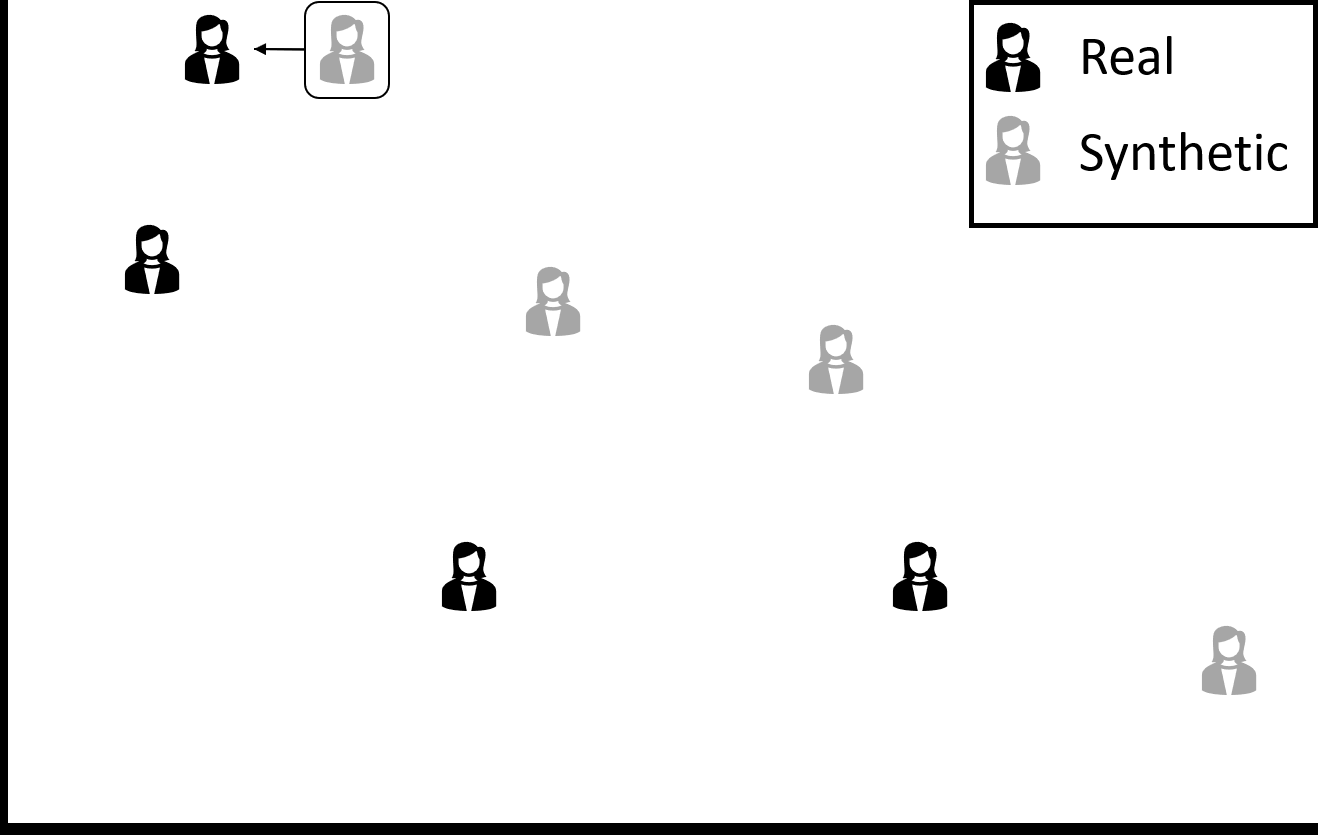}
         \caption{Smallest synthetic to real distance (SRD) of selected synthetic record}
    \end{subfigure}\hfill
    \begin{subfigure}{0.46\linewidth}
        \centering
        \includegraphics[width=\linewidth]{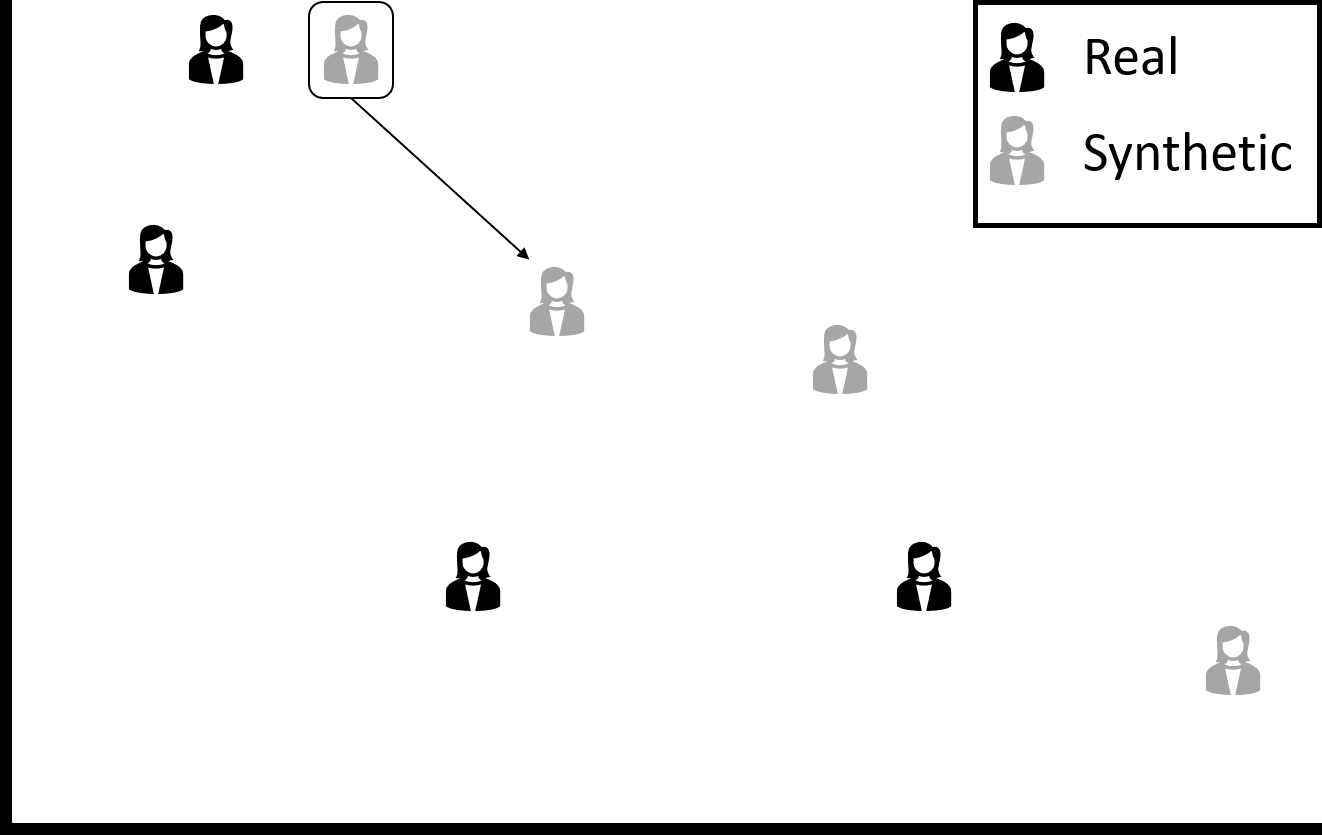}
        \caption{Smallest synthetic to synthetic distance (SSD) of selected synthetic record}
    \end{subfigure}\\

    \begin{subfigure}{0.46\linewidth}
        \centering
         \includegraphics[width=\linewidth]{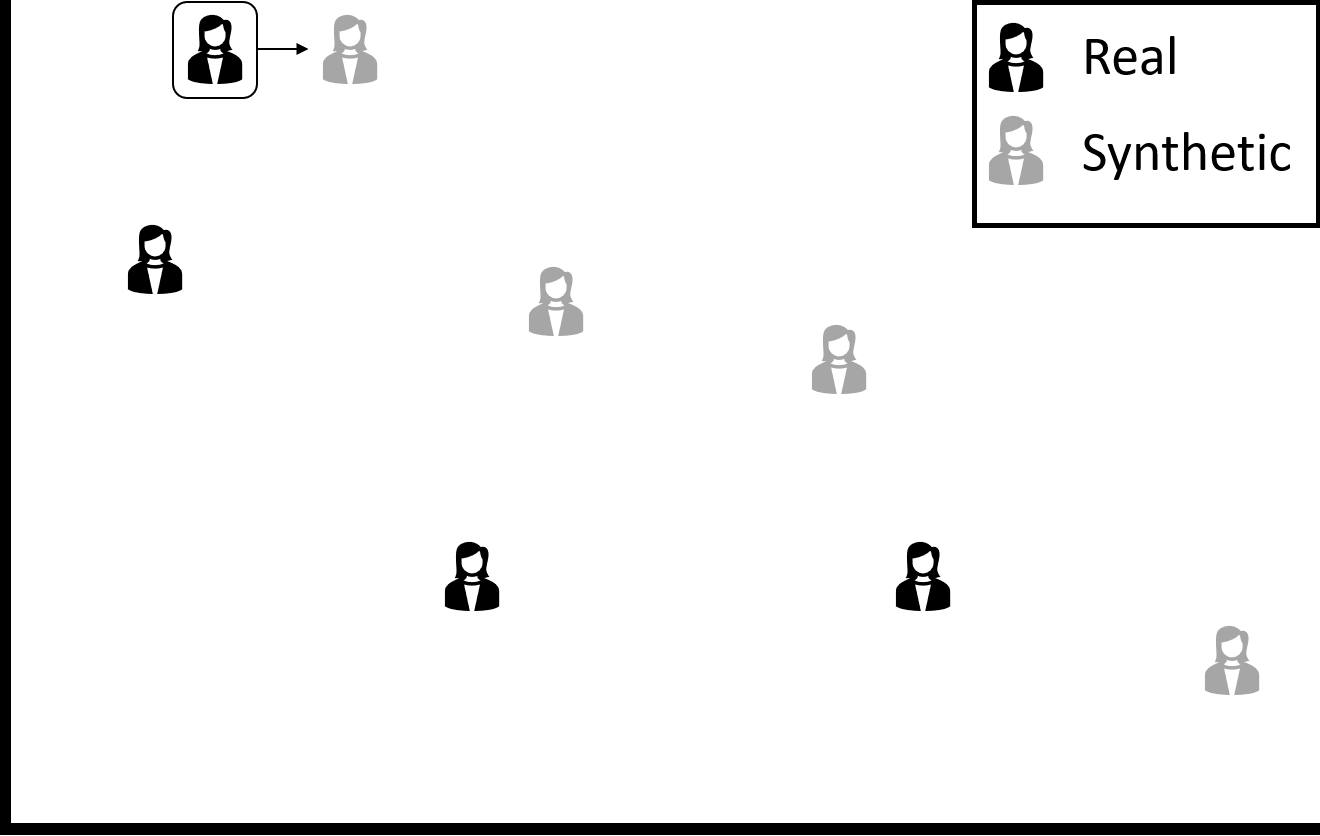}
         \caption{Smallest real to synthetic distance (RSD) of selected real record}
    \end{subfigure}\hfill
    \begin{subfigure}{0.46\linewidth}
        \centering
        \includegraphics[width=\linewidth]{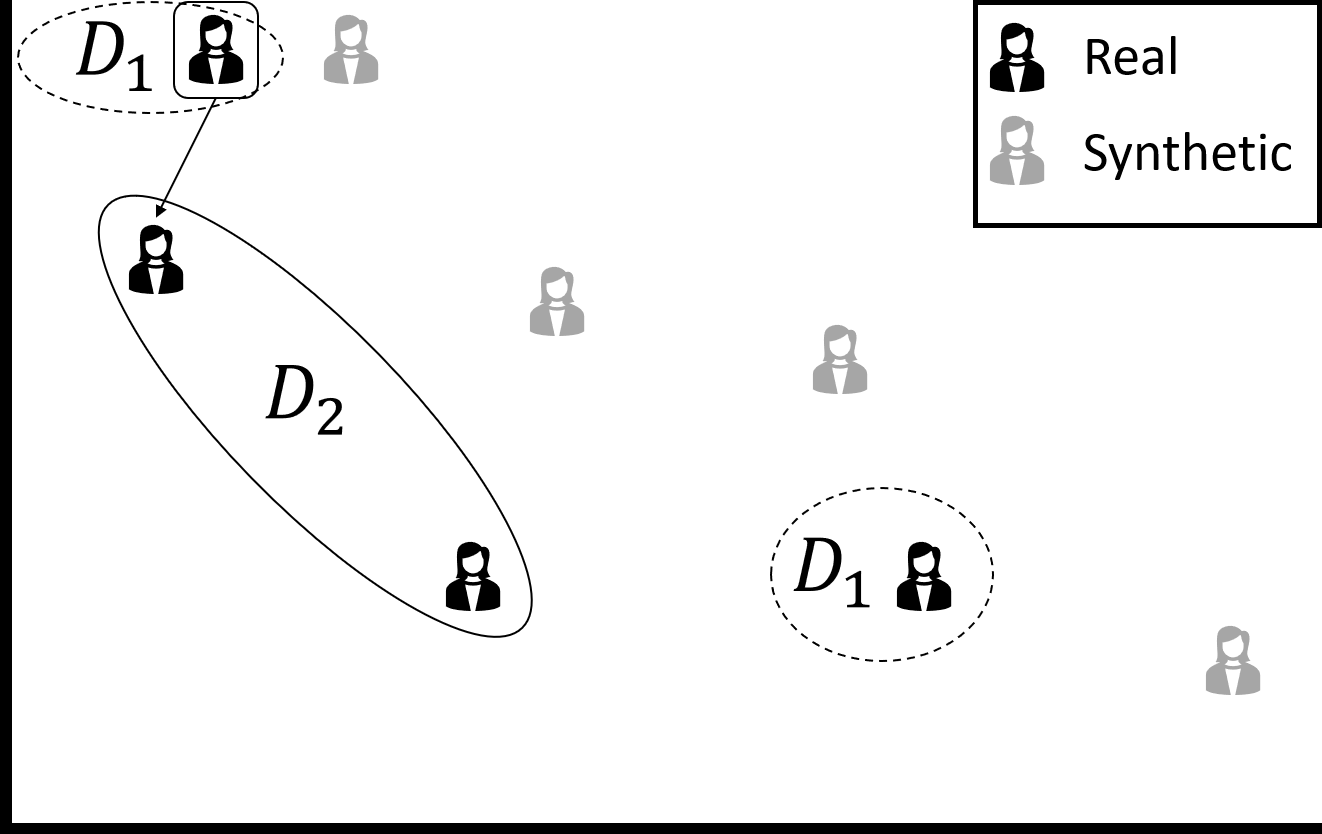}
        \caption{Smallest real to real distance (RRD) of selected real record}
    \end{subfigure}\hfill
    
    \caption{Distances evaluated for given synthetic and given real records in privacy indicators. Most indicators involve computing at least SR and RR distances for all synthetic and real data points. In (d), the record is in $D_1$ and is therefore only compared to real records in $D_2$.}
    \label{fig:dist}
\end{figure}

        \begin{figure}[H]
       \centering
        \includegraphics[width=0.89\textwidth]{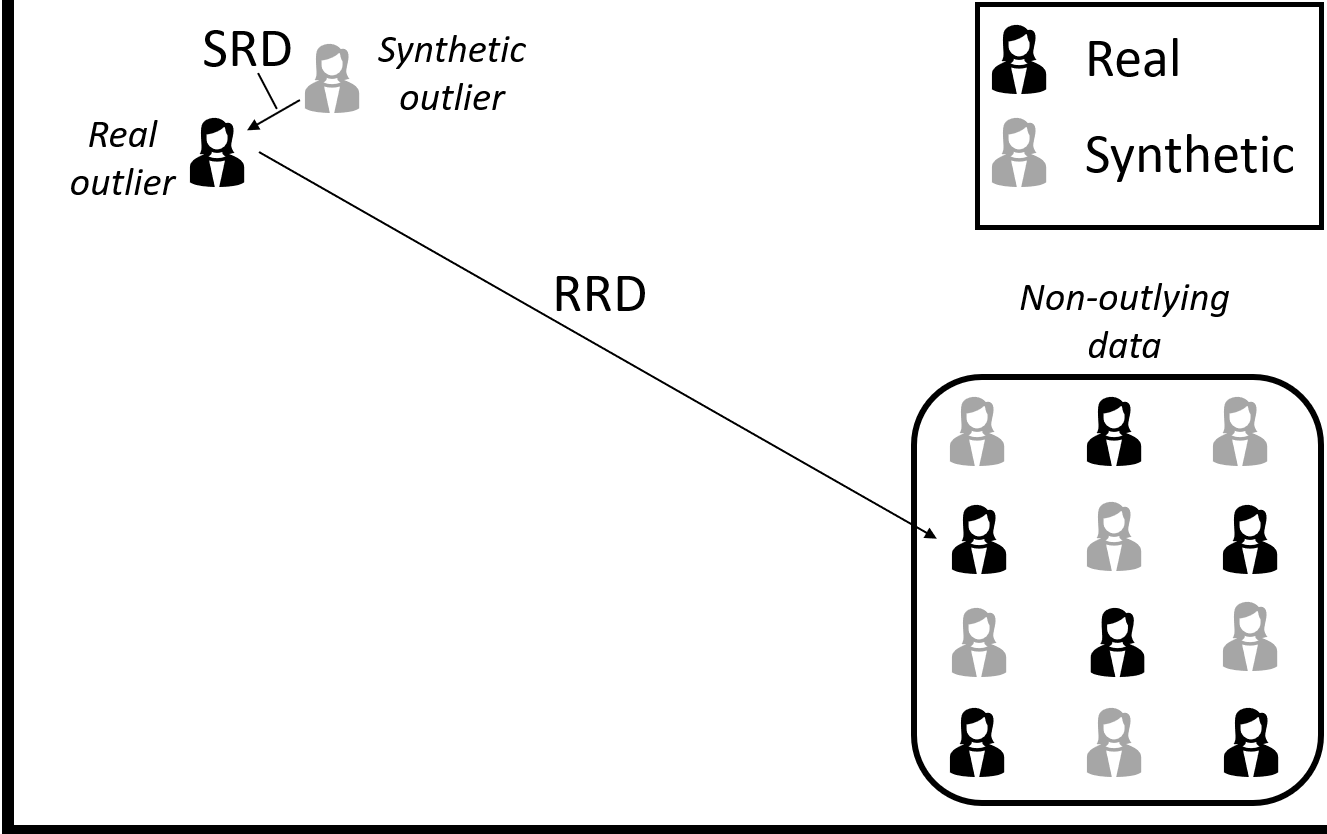}
        \caption{The DCR is sensitive to outliers: if the SD is accurate, it reproduces an outlier similar to the real outlier. The distance between the synthetic and real outliers (SRD) is then small. By definition, the distance between the real outlier and the closest other real record (RRD) is relatively large. Thus, $\texttt{SRD}(\hat{d})<\texttt{RRD}(d^*)$.}
        \label{fig:SRD}
    \end{figure}

    \begin{table}[H]
    \centering
    \resizebox{\textwidth}{!}{%
    \begin{tabular}{lllllllll}
        \toprule
                       &             & \textbf{Distance}  & Similarity & Evaluation &             &          &             &\\
       \textbf{Method} & \textbf{IMS}&\textbf{based}      & metric     & method     & Statistic(s)  & Holdout? & \textbf{NN} & \textbf{Other}\\
       \midrule
       \cite{Emam}& - & -         & -         & -    &-         & -        & -           & Statistics \\
      \cite{esteban2017realvalued}& -& - & - & -& -& -& - & MMD \\
       \cite{smooth}& -& - & -& -& -& -& - & MMD \\
      \cite{task}& Yes & -         & -         & -        &-     & -        & -           & - \\
      \cite{NIST}& Yes & -         & -         & -        &-     & -        & -           & - \\
       \cite{Raab} & Yes & -         & -         & -   &-          & -        & -           & - \\
     \cite{VENUGOPAL2022339}& - & PL   & Euclidean  & NS    & PL         & Yes                & -           & - \\
       \cite{Yale3}& - & AA; PL       & NS         & NS   & AA; PL &Yes     & -           & - \\

       \cite{zhao2021ctabgan}& - &DCR & Euclidean & NS & Percentiles & No & Yes & -\\
       \cite{Ebert}& Yes & DCR & NS & NS & Percentiles & NS & Yes & -\\
      \cite{LuHitting} & Yes & DCR & Euclidean & NS & $\mu,\sigma$ & No & - & -\\
       
       \cite{language} & - & DCR & $\mathcal{L}_1$ (num) & Aggr. & Histogram & No & - & -\\
                                                  &   &     &~~Hamming (cat) &&  &  &  &  \\
       \cite{solatorio} & - & DCR & NS &NS & Histogram & Yes & - & -\\
       \cite{an2023distributional}&- & DCR & Euclidean & Ign. & $\mu,\sigma$&No & - & -\\
       \cite{Diffusion} & - & DCR & Euclidean & NS& Median & No & - & - \\

       \cite{kotelnikov2022tabddpm}& - & DCR & NS &NS &Median & No & - &- \\
       \cite{5g}& Yes & DCR & NS & NS & Percentile & No & - & - \\
       \cite{zhang2023generative} & - & DCR & $\mathcal{L}_1$ (num) + & Aggr. & Histogram & No & - & -\\
                                                 &   &     &~~Hamming (cat) &&  &  &  &  \\
       \cite{Kunar} & - & DCR & Euclidean & NS & Percentile & No & Yes & - \\

       \cite{Norgaard} & - & DCR & Cosine & NA & $\mu$ & yes & - & -\\
       \cite{weldon2021generation}&Yes & DCR & Euclidean & NS & $\mu,\sigma$& Yes & - & -\\
       \cite{Platzer}& Yes & DCR & Hamming & Bin. & $p,\mu$ & Yes & Yes & -\\
       \cite{NeurIPS}&- & DCR & Euclidean & Emb. & Percentile & Yes & - & -\\

       \cite{PanfIEEE, PhD_Dani}&- & DCR & Euclidean& Emb. & $p,\mu,\sigma$ & Yes & - & - \\
                                                  &  &     &        &&~~Inferential$^*$&     &   &   \\
       \cite{HERRANZ201578} & - & DBRL & Euclidean & NS & $p$ & No & - & -\\
       \cite{DBRLSynth}& - & DBRL & Euclidean; & NS & $p$ & No & - & -\\
         &  &  &~~Manhalanobis & && &  & \\

         \cite{avatar} & Yes & DCR & Euclidean &Emb. &Percentile & No & Yes & Seed\\
         
        \bottomrule
    \end{tabular}%
    }
    \caption{Indicators used in practice; ``NS'': Not specified; ``num'': for numeric attributes; ``cat'': for categorical attributes; ``Aggr.'': aggregating two metrics (one for numerical and one for categorical attributes); ``Ign.'': ignoring categorical attributes; ``Emb.'': evaluating indicators in an embedding space; ``bin.'': binning numeric attributes; ``NA'': Not applicable, for instance because data types are not mixed in the dataset(s) of the involved study; $p$: proportion; $\mu$: mean; $\sigma$: standard deviation; seed: seed-specific distance-based indicators other than DBRL, e.g. local cloaking; hidden rate. *: the authors use a Kolmogorov-Smirnov test to test the null-hyopthesis: ``the SRD and RRD distributions stem from the same underlying distribution'', with $\alpha$ levels of 0.05 and 0.01.}
    \label{tab:indicators}
\end{table}

\section{Empirical Privacy Assessment Frameworks}\label{app:attack}

    \begin{figure}[H]
       \centering
        \includegraphics[height=0.25\textheight]{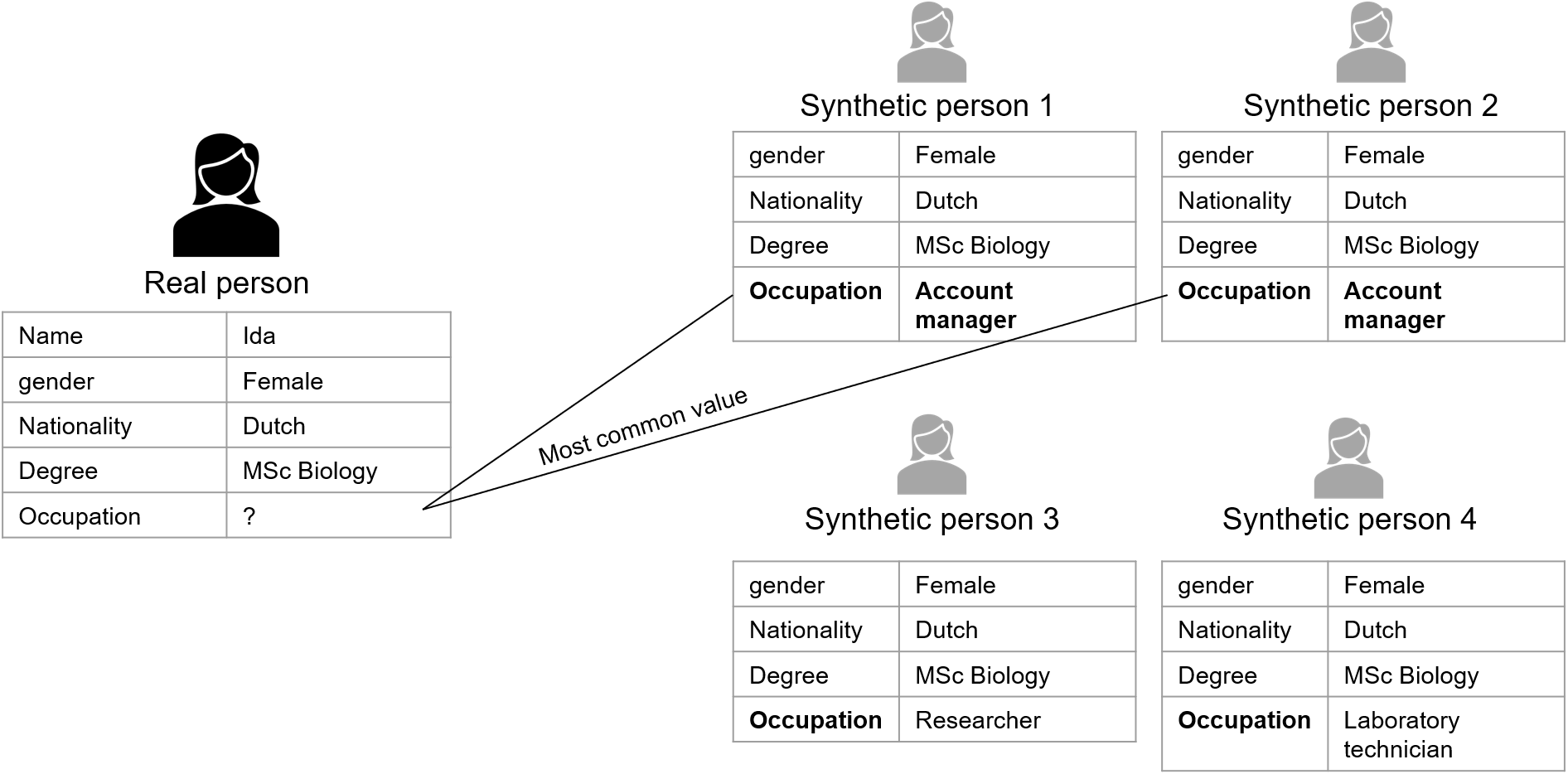}
        \caption{Nearest neighbor attack based on nearest neighbors (NN). The adversary knows Ida's gender, nationality and degree, but not her occupation. The adversary has access to a synthetic dataset. In this synthetic set, the adversary searches for the $k$ (in this case four) records with the most similar profile based on the known attributes. The adversary then infers Ida Jansen's occupation based on the $k$ closest synthetic records. The synthetic record's values can be aggregated in a number of ways. E.g. Ida can be assigned the most common value of the neighbors (in this case: account manager). For numeric attributes, averages can also be taken.}
        \label{fig:NNattack}
    \end{figure}

    \begin{figure}[H]
       \centering
        \includegraphics[height=0.18\textheight]{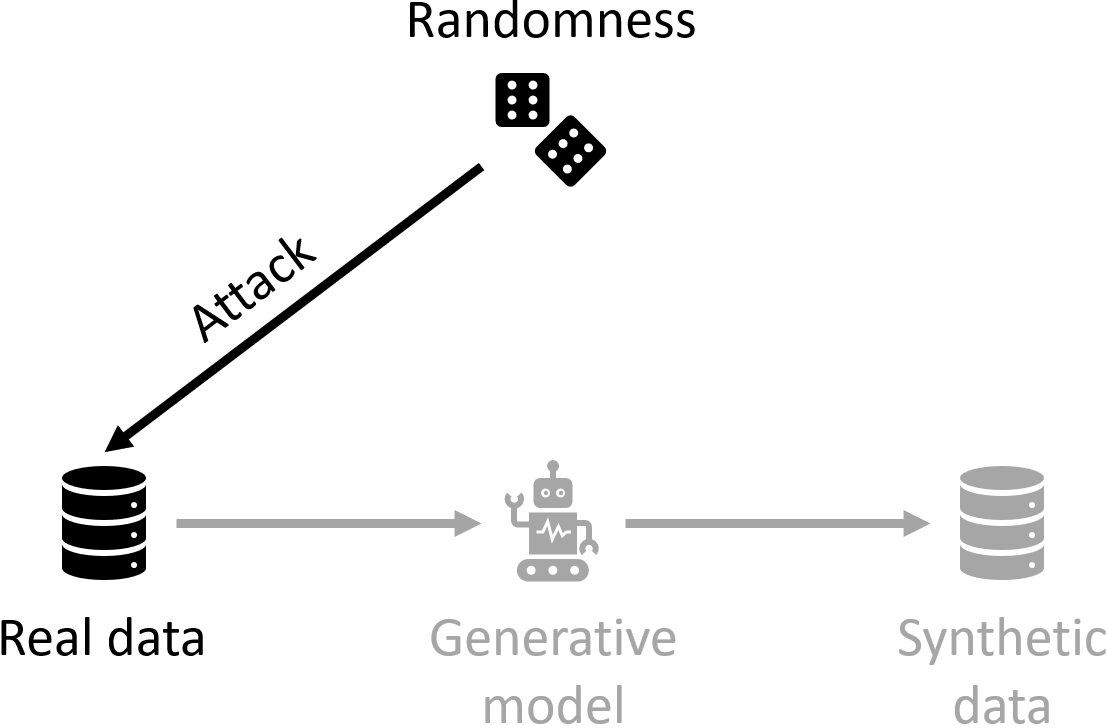}
        \caption{Random baseline to benchmark successful attacks against (cf. Figure~\ref{fig:base1})}
        \label{fig:base2}
    \end{figure}

    \begin{table}[ht]
    \centering
    \resizebox{\textwidth}{!}{%
    \begin{tabular}{llllll}
        \toprule
                                                                  &  Threat   &          & Attack    & Attack     & Attack      \\
       Method                                                     &  model    & Baseline & estimator & technique  & type (WP29) \\
       \midrule
       \cite{carlini2019secret}                  & No box    & SL (IT)  & IT        & VRD        & S\\
       \cite{narayanan2006break}& Aux       & A        & IT        & VRD        & L     \\
       \cite{choi2018generating}    & Aux       & SL       & NN        & VRD        & I$^*$\\
       \cite{membership}                      & Black; aux& SL (PR)  & ML        & SM; MIA    & S, I$^*$ \\
       \cite{knock} & Aux       & SL       & ML        & MIA        & S, I$^*$ \\
       
       \cite{Anonymeter}    &           &          &           &            &             \\
       ~~\emph{Singling out}                                      & No box    & R; G     & -         & VRD        & S\\
       ~~\emph{Linkage}                                           & Aux       & R; G     & NN        & VRD        & L     \\
       ~~\emph{Inference}                                         & Aux       & R; G     & NN        & VRD        & I$^*$\\

       \cite{TAPAS}  &           &          &           &            &              \\
       ~~\emph{Neighborhood}                                      & Aux       &  R       &  NN       & VRD        &  I$^*$   \\
       ~~\emph{Inference}                                         & Aux       &  R       &  ML       & VRD        &  I$^*$   \\
       ~~\emph{Shadow model}                                      & Black     &  R       &  ML       & SM; MIA    &  S, L, I         \\

      \cite{Goncalves}             &           &          &           &            &              \\
      ~~\emph{Attribute inf.}                                    & Aux       &  SL      &  NN       &  VRD       &  I$^*$\\
       ~~\emph{Membership inf.}                                   & No box    &  SL      &  NN       &  MIA       &  S, I   \\
       \cite{Yale3}           & Black; aux& SL       &  NN       & MIA        & S$^*$\\
      \cite{groundhog}             & Black; aux& M (DP)   &  ML       & MIA        & L       \\
       \cite{Park_2018}              & White; aux& SL (AUC,F1)& ML       & SM; MIA    & S$^*$\\
      \cite{vanbreugel2023membership}    & Black; aux& SL (AUC)  & NN; ML   & SM; MIA    & L       \\

       \cite{Kuppa}                   & Black     & SL (F1)   & ML       & SM; MIA    & S  \\
       \cite{oprisanu2022utility}      &           &          &           &            &              \\
       ~~\emph{Limited aux}                                       & Aux       & R        & ML        & SM; MIA    & S$^*$\\
       ~~\emph{Aux}                                               & Aux       & R        & ML        & SM; MIA    &  L     \\
       \cite{meeus2023achilles} & Black    & SL (AUC) & ML        & VRD; MIA    & S; I\\
       \cite{Diffusion} & No box & A & NN & VRD; MIA & I\\
         
        \bottomrule
    \end{tabular}%
    }
    \caption{Empirical privacy assessment frameworks. Threat model: aux - auxiliary information; black, white - black box, white box; Mult: experiments conducted with multiple threat models; Baseline: A - absolute (proportion or quantity of correct hypotheses, etc.), M - privacy metric ($k$ - $k$-anonimity, DP - differential privacy), R - random, C - control, SL - metrics from supervised learning (PR - precision and recall, AUC - area under ROC curve, F1 - F1-score); Attack estimator: IT - information theory, NN - nearest neighbor, ML - machine learning; Attack technique: VRD - vulnerable record discovery through sorting, searching or sampling, SM - shadow modeling, MIA - membership inference attack; Attack type (WP29): S - singling out; L - linkage; I - inference, $^*$: technically, any attack using auxiliary information is a linkage attack to some degree.}
    \label{tab:attack_comp}
\end{table}
\end{document}